\documentclass[letterpaper, 10 pt, conference]{ieeeconf}
\IEEEoverridecommandlockouts
\let\labelindent\relax
\usepackage{graphicx}
\usepackage{subcaption}
\usepackage{eqlist}
\usepackage{amsfonts}
\usepackage{tabularx}
\usepackage{booktabs}
\usepackage{amssymb}
\usepackage{amsmath}
\usepackage{enumitem}
\usepackage{threeparttable}
\usepackage[lined, ruled, linesnumbered, commentsnumbered, noend]{algorithm2e}
\usepackage{lipsum}

\usepackage[usenames,dvipsnames]{xcolor}
\usepackage{comment}
\usepackage{wrapfig}

\definecolor{cpurple}{rgb}{0.93,0.098,0.584}
\usepackage{multirow}

\usepackage{url}
\usepackage{CJKutf8}
\definecolor{bleudefrance}{rgb}{0.19, 0.55, 0.91}

\begin{document}
\title{\LARGE \bf Metal Wire Manipulation Planning for 3D Curving --- How a Low Payload Robot Can Use a Bending Machine to Bend Stiff Metal Wire}

\author{Ruishuang Liu$^{1}$, Weiwei Wan$^{1*}$, Emiko Isomura$^{2}$, and Kensuke Harada$^{13}$% <-this % stops a space
\thanks{$^{1}$Department of System Innovation, Graduate School of Engineering Science, Osaka University, Toyonaka, Osaka, Japan.}
\thanks{$^{2}$First Department of Oral and Maxillofacial Surgery, Graduate School of Dentistry, Osaka University, Japan.}
\thanks{$^{3}$National Inst. of AIST, Japan.}
\thanks{$^{*}$Contact: Weiwei Wan, {\tt\small wan@sys.es.osaka-u.ac.jp}}}

% \markboth{IEEE Transactions on Automation Science and Engineering. Submission for Review, 2022}
% {Liu \MakeLowercase{\textit{et al.}}: Title}

\maketitle

\begin{abstract}
This paper presents a combined task and motion planner for a robot arm to carry out 3D metal wire curving tasks by collaborating with a bending machine. We assume a collaborative robot that is safe to work in a human environment but has a weak payload to bend objects with large stiffness, and developed a combined planner for the robot to use a bending machine. Our method converts a 3D curve to a bending set and generates the feasible bending sequence, machine usage, robotic grasp poses, and pick-and-place arm motion considering the combined task and motion level constraints. Compared with previous deformable linear object shaping work that relied on forces provided by robotic arms, the proposed method is suitable for the material with high stiffness. We evaluate the system using different tasks. The results show that the proposed system is flexible and robust to generate robotic motion to corporate with the designed bending machine.
\end{abstract}

% \begin{IEEEkeywords}
% Manpulation Planning
% \end{IEEEkeywords}

\section{Introduction}
\label{sec:introduction}
This paper presents a combined task and motion planning method for a robot to curve metal wires into 3D shapes. We assume a collaborative robot that is safe to work in a human environment. Such a robot cannot provide enough force to bend objects with large stiffness. To solve the problem, we designed a bending machine using a stepper motion and a 1:80 gear set. The bending machine can provide 240Nm bending force to bend stiff objects. We develop a combined task and motion planner to plan the bending sequence of the bending machine at the task level and the robot's grasping poses and joint trajectories at the motion level. The planner enables the robot to use the bending machine like a professional human worker and curve stiff and various-shaped metal wires without manually defined rules or programming routines.

Previously, using robots to curve metal wires was widely studied, as robots provided a promising solution to flexible or personalized manufacturing of metal-wire products. Existing robotic methods use robots as independent machines. They program a robot to hold a metal wire and bend it by pressing against a fixture or an external robotic gripper. The payload of the robot limits the maximally affordable metal stiffness. Unlike the existing methods, we ask a robot to collaborate with an external folding machine. With the support of our proposed combined task and motion planning method, the robot can automatically determine the folding sequence, the grasping poses, and the robot joint trajectories to conduct 3D curving of stiff metal wires.

\begin{figure}[!tpb]
  \begin{center}
  \includegraphics[width=\linewidth]{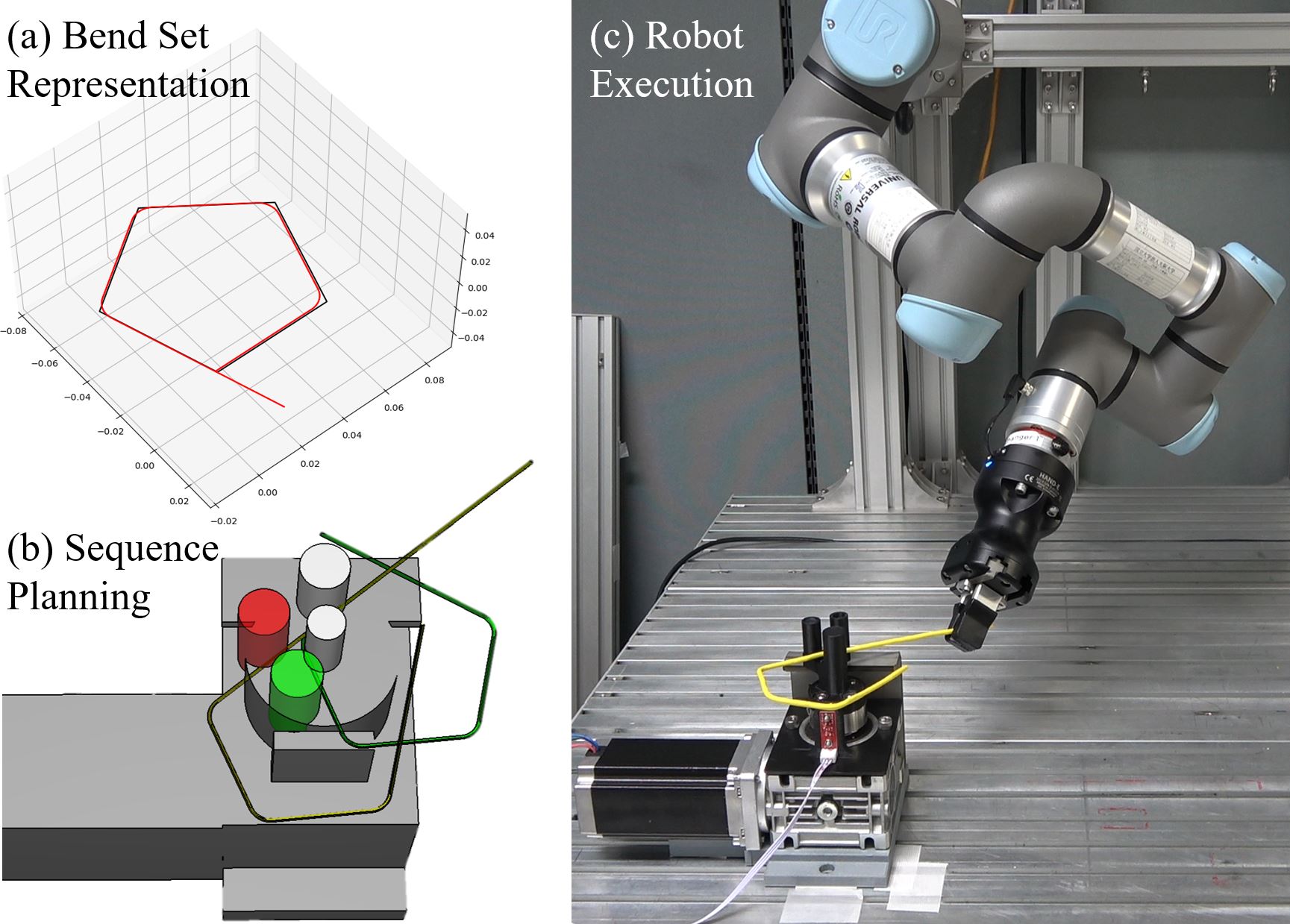}
  \caption{(a) A desired curve is represented as a bending set using polygonal approximation. (b) Close-loop planning considering combined task and motion level constraints. (c) A collaborative robot (low payload) bends a stiff metal wire using a bending machine while following the planned result.}
  \label{fig:teaser}
  \end{center}
\end{figure}

Fig. \ref{fig:teaser} shows an example of our system in action. In the first step, we use polygonal approximation to represent a desired 3D curve as a set of bending candidates. Then, we perform a combined task and motion planning to simultaneously generate the bending sequence and machine usage, the grasping poses, and the pick-and-place motion of the robot. The robot will follow the planned results to use the bending machine and bend metal wire.

Our main contribution is the close-loop exploration considering combined task and motion level constraints. At the task level, we study the geometric constraints of the bending machine and the changing shapes of the metal wire. These constraints help us to determine several tentative bending sequences. At the motion level, we study the logical relations of robotic grasping poses and the availability of robot joint trajectories that move the metal wire from a previous bending pose to the next one. The motion level results will be recorded during exploration to prune the task level tentative sequences and improve planning speed. They together make the impossible exploded combinatorics tractable.

In the experimental section, we carry out both simulations and real-world experiments to study the performance of the proposed method. The results show that our method can find a solution for metal wire with less than eight bending candidates in two minutes if a solution exists. The developed robot system can curve different 3D shapes in the real world with satisfying performance.

% The remaining part of this paper is organized as follows. Section \ref{sec:relatedwork} presents the related studies. Section \ref{sec:method} presents the preliminary knowledge and definitions of notations. Section \ref{} dives into the combined task and motion planning and pruning algorithms. Section \ref{} presents the experiments and analysis. Conclusions are drawn in Section \ref{}.

\section{Related Work}
\label{sec:relatedwork}
\subsection{Manipulation of Deformable Linear Objects}
Robotics research communities have paid considerable attention to the problem of the automatic manipulation and shaping of deformable linear objects. The objective is to reshape an object into the desired one using a mechanical system. Different from a rigid object, a deformable object is more complicated due to its infinite degree of freedom. The large configuration space of deformable objects makes it difficult to model using traditional approaches \cite{arriola2020modeling}. Also, since the object shape changes while being manipulated, additional methods need to be developed to track the deformation and conduct online collision checking during planning. 

State-of-the-art deformable objects studies in robotic manipulation literature can be separated according to their physical properties (with and without compression strength) and geometric properties (linear, planar, and solid) \cite{sanchez2018robotic}.
Especially for DLOs, they are further classified into five categories in \cite{henrich1999manipulating} considering its deformable type (plastic or elastic) and whether force less than gravity can result in deformation.

This paper focuses on elastoplastic DLOs (e.g. metal wire, rod). Examplary studies on this category are as follows. Jin et al. \cite{jin2013bending} designed an automation device to perform the archwire bending task. The device could form several desired shapes by predefined processes. Xia et al. \cite{xia2016development} achieved the same task with a robotic arm and an external gripping unit. A sample-based planner is applied for bending path generation. More recently, Laezza et al. \cite{laezza2021learning} proposed a reinforcement learning-based elastoplastic DLOs shape control method. When the material's stiffness is high, it will be difficult for general collaborative robots to reshape. Solutions include Lu et al. \cite{lu2020transient}, which leveraged a heating technique in cooperation with an industrial robotic arm to soften shape metal rods, and Zhang et al. \cite{zhang2022design}, which designed a special multi-axis special-purpose machine to form titanium alloy strip used in oral and maxillofacial surgery.

Unlike the studies above, we propose using a collaborative robot arm and a bending machine to achieve the curving task. Our difference and novelty are two-fold. First, the proposed system can cope with high stiffness metal wire, as bending forces do not directly affect the robot arm. Second, as the metal wire is picked and held by the robot arm rather than fixed on the machine or special end-effectors, diverse bending sequences are allowed to support combined task and motion planning.

\subsection{Combined Task and Motion Planning}
Combined exploration and planning considering the mutual influence of task and motion level constraints is the key technique for efficiently generating robotic manipulation motion using the bending machine and bending metal wire. The method extends combined task and motion planning, which was used to generate action sequences at a high level and motion for specific actions at a low level. While combined task and motion planning has been conventionally studied for decades, recent literature attempts to extend it with uncertainty, force constraints, and deeply learned heuristics. For example, Holladay et al. \cite{holladay2021planning} extended an existing task and motion planner with controllers that exert wrenches and constraints while considering torque and frictional limits to achieve Forceful Manipulation. Silver et al. \cite{silver2021learning} proposed a bottom-up relational learning method for operator learning and demonstrated how the learned operators could be used for planning in a TAMP system. Previously in our group, we also adopted the idea of combined task and motion planning for various robotic manipulation tasks. 
Chen et al. \cite{chen2021} developed a planner that automatically finds an optimal assembly sequence for a dual-arm robot to build a woodblock structure while considering various constraints and supporting grasps from a second hand. 
Wan et al. \cite{wan2022arranging} presented a combined planner to solve a test tube arrangement problem. 

In this work, we use similar combined task and motion planning methods to get bending sequence considering constraints of the bending machine, reason the grasp poses for metal wire, and generate manipulation motion as Chen et al. \cite{chen2021}. We pay special attention to the geometric constraints of the bending machine, the changing shapes of the metal wire, and the kinematic constraints of the robot. We determine tentative bending sequences and prune them considering the mutual influences of the constraints. 

\section{Preliminaries}
\label{sec:method}

\subsection{Structure of the Bending Machine}
\begin{figure}[htb]
    \centering
    \includegraphics[width=\linewidth]{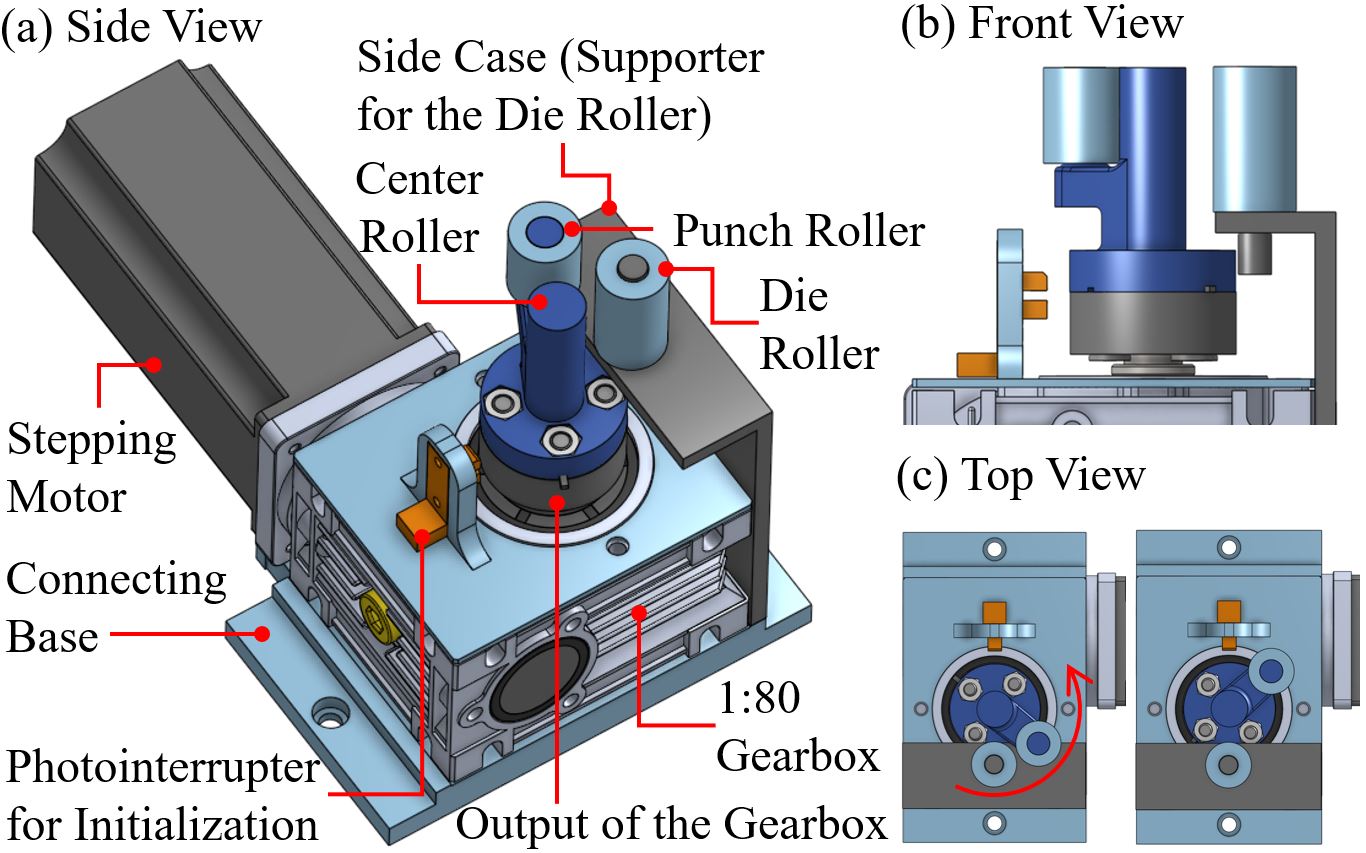}
    \caption{CAD model of the designed bending machine.}
    \label{fig:folding_machine}
\end{figure}
\begin{figure}[htb]
    \centering
    \includegraphics[width=8.4cm]{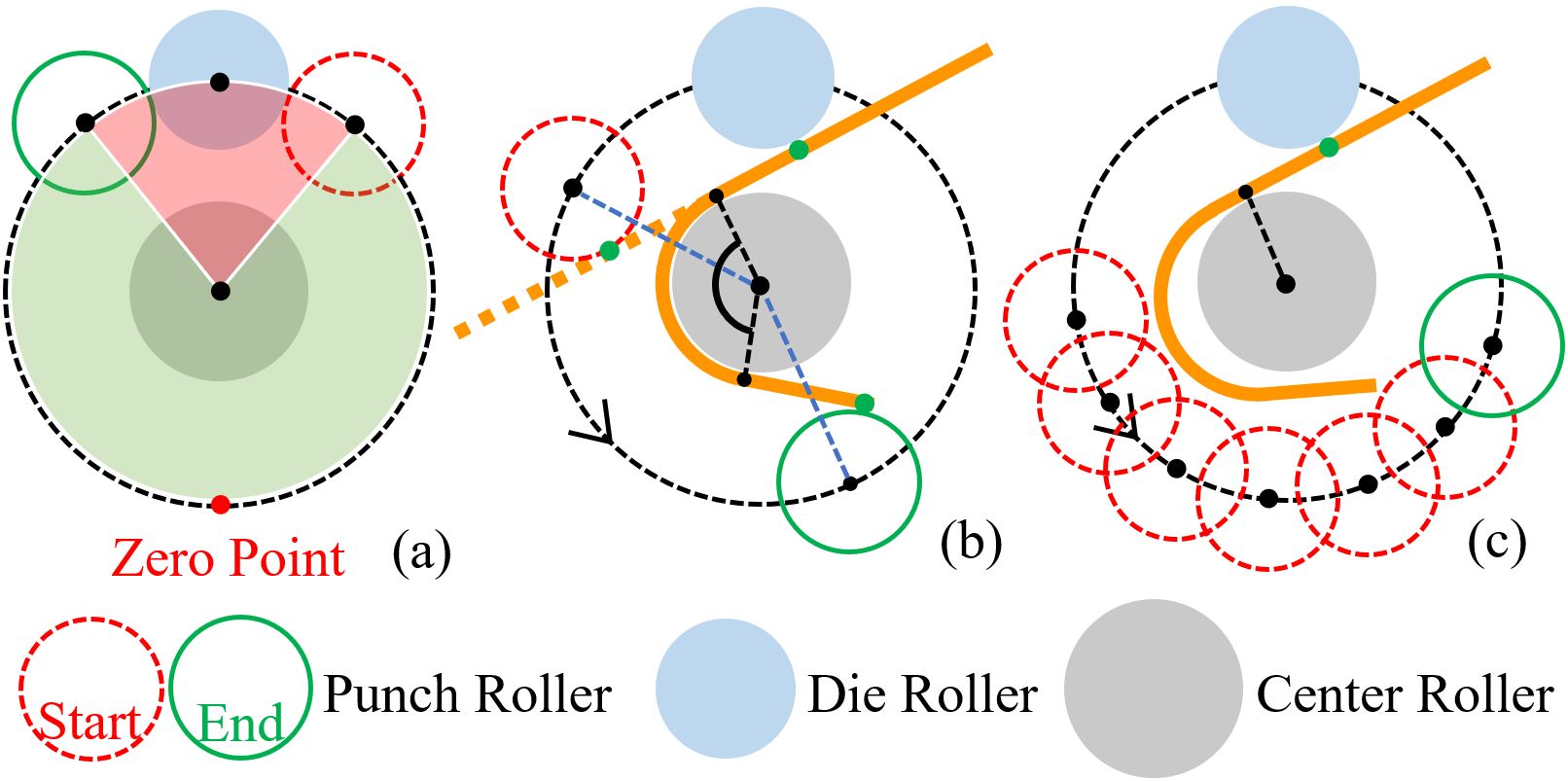}
    \caption{(a) Effective work range of the bending machine assuming zero-thickness metal wire. (b) One wire shape pressed by the punch roller. (c) If the punch roller cannot touch the metal wire, the bending action is reported as infeasible. For simplicity, the figures are illustrated in a 2D top view. They are actually three dimensional as shown in Fig. \ref{fig:bending_representation} and \ref{fig:angles}.}
    \label{fig:working_range}
\end{figure}

We designed a bending machine shown in Fig. \ref{fig:folding_machine} by following the mechanism of commercial metal plate benders or folders. The folding machine comprises a 3Nm stepping motor, a 1:80 gearbox to increase the output torque, an optical sensor as a photo-interrupter for calibrating a zero position, and several rollers for pressing metal wires. A metal wire is assumed to be placed between the center roller and the other two rollers. The punch roller can rotate clockwise and counterclockwise by starting from the calibrated zero position. Thus the machine allows bending from two directions without reversing the plate. The effective work range of the machine is shown in Fig. \ref{fig:working_range}(a). Its maximum bending angle for a zero-thickness metal wire will be the green zone of the figure.

The actual bending angle depends on the shape and thickness of the metal wire. It is essentially smaller than the maximum bending angle. In an ideal case where the metal wire is straight, the bending angle can be computed directly considering the punch roller's rotating angle and wire thickness. It is difficult to figure out an explicit equation to formulate the bending in more common cases. Thus instead of modeling the bending process, we develop a kinematic simulator to examine the collision between the wire and the environment (including robots, other parts of the bending machine, work table, etc.) and determine the next curve shape after the current bending action.

We have the following assumptions for the kinematic simulator. First, we assume the bending force from the bending machine is large enough to cause inelastic deformation. No kinetic energy will be accumulated and released to cause unexpected results. Second, the wire begins to bend when the punch roller starts to collide with it and stops bending when the punch roller reaches a limit, or the wire collides with the environment. Third, we assume the bending arc is clinging to the center roller during bending. The wire will wind around the center roller to form a circular arc. Fourth, if two bending actions overlapped, we assume the newer circular arc overrides the former arc sections. Under these assumptions, we can compute the next shapes after each bending by replacing related wire sections with circular arcs determined by collisions. Note that although we use this design to present our combined task and motion planning implementation, our method is not limited to this specific machine. It can be applied to other bending mechanisms \cite{ito2016collaborative}\cite{hamid20163d}.

\subsection{Representing a 3D Curve Using a Bending Set}

We formulate the bending problem considering the bending machine as a searching problem. We simplify a given desired 3D curve using polygonal approximation as a first step. We assume a desired 3D curve is initially given as a list of dense points. This representation includes many delicate curvatures that our bending machine cannot bend. Instead of directly using these dense points, we approximate them by line segments using the Ramer-Douglas-Peucker algorithm \cite{ramer1972iterative}, while considering an error tolerance $\varepsilon$.

\begin{figure}[htb]
    \centering
    \includegraphics[width=\linewidth]{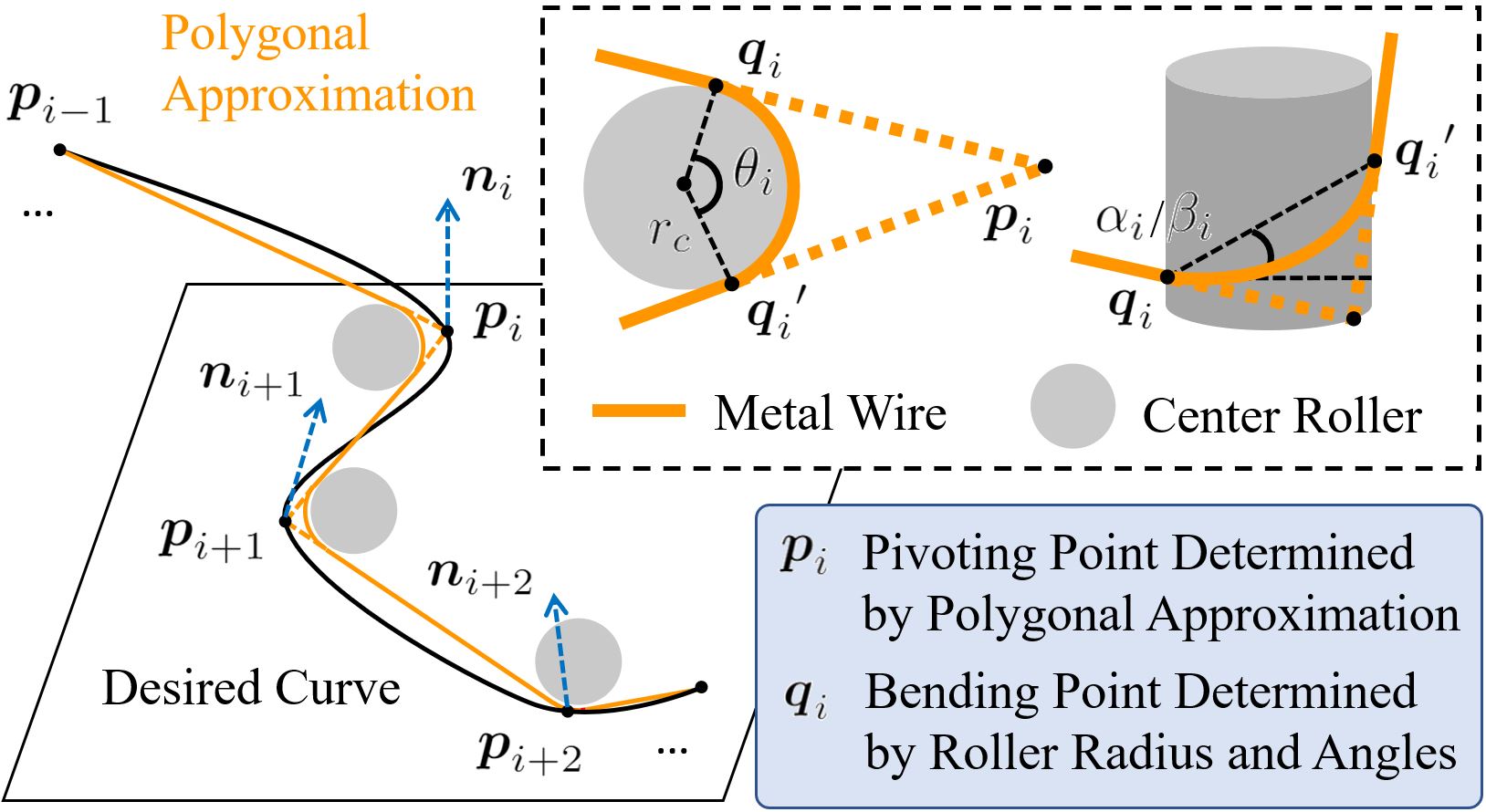}
    \caption{Polygonal approximation of a desired curve. The polygonal approximation leads to a set of bending candidates defined as $\{q_i, \theta_i, \alpha_i, \beta_i\}$. They are illustrated in the dashed box. Please also see Fig. \ref{fig:angles} for details of the $\theta_i, \alpha_i, \beta_i$ angles.}
    \label{fig:bending_representation}
\end{figure}

\begin{figure*}[!tpb]
  \begin{center}
  \includegraphics[width=.97\linewidth]{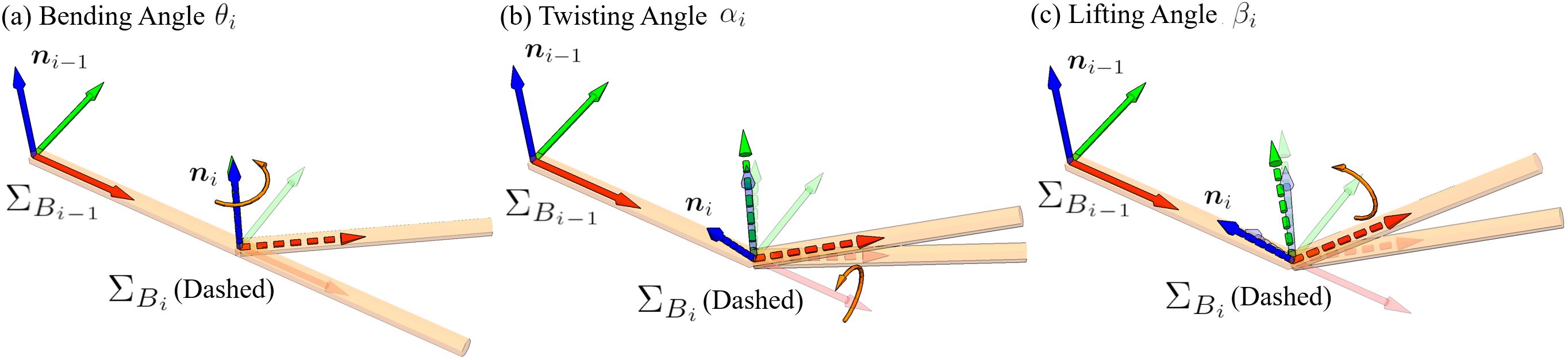}
  \caption{Three angles used to represent a bending state. They are defined in the local coordiante system of a previous bend. Since the metal wire is central symmetric, using both $\alpha$ and $\beta$ are redundant. The combined task and motion planner will determine which one to use considering environmental constraints.}
  \label{fig:angles}
  \end{center}
\end{figure*}

Fig. \ref{fig:bending_representation} illustrates a example of polygonal representation. Here, the smooth black curve is the originally desired one. The orange polygonal curve is the approximated result. The approximation results into several pivoting points named $\boldsymbol{p}_{i-1}$, $\boldsymbol{p}_i$, $\boldsymbol{p}_{i+1}$, ... and their normals $\boldsymbol{n}_{i-1}$, $\boldsymbol{n}_i$, $\boldsymbol{n}_{i+1}$, ..., since the curve is a spatial one. By further considering the radius of the center roller $r_c$, we can compute a bending set $\mathbb{B}=\{B_i\}$ where each element in it indicates a bending candidate represented as $B_i=\{q_i, \theta_i, \alpha_i, \beta_i\}$. Here, $q_i$, $\theta_i$, $\alpha_i$, and $\beta_i$ are respectively the starting point of the bending, the wire's bending angle, the wire's twisting angle, and the wire's lifting angle at this point. These parameters are described in the local coordinate system of its preceding bending point $B_{i-1}$, as shown by Fig. \ref{fig:angles}. The definition of $q_i$ is graphically illustrated in the dashed box of Fig. \ref{fig:bending_representation}. The metal wire begins to wind around the center roller starting from this point. The three angles are illustrated in both the dashed box of Fig. \ref{fig:bending_representation} and Fig. \ref{fig:angles}. They represent how to rotate the metal wire to reach a desired local shape.

Note that since the metal wire is a cylinder and central symmetric, a local shape essentially has two degrees of freedom and can be defined by two parameters. Thus it may seem unnecessary to use both the $\alpha$ and $\beta$ angles to represent the $B_i$. This understanding is true from the viewpoint of the final shape, but choosing $\alpha$ or $\beta$ will lead to different wire motions during bending. The different wire motions could be helpful to avoid collisions. Thus, instead of using a single angle parameter, we use both of them $B_i$'s definition. Considering environmental constraints, the combined task and motion planner will determine which one to use.

Using the radius of the center roller $r_c$, the values obtained from polygonal approximation, i.e. $\boldsymbol{p}_{i-1}$, $\boldsymbol{p}_i$, ..., and their normals $\boldsymbol{n}_{i-1}$, $\boldsymbol{n}_i$, ..., the parameters of $B_i$ can be computed as follows.
\begin{align}
    \textnormal{Pj}_{\Sigma_{B_{i-1}}^{(x y)}}(\overrightarrow{\boldsymbol{p}_{i}\boldsymbol{p}_{i+1}})=\overrightarrow{\boldsymbol{p}_{i}\boldsymbol{p}_{i+1}}-\overrightarrow{\boldsymbol{p}_{i}\boldsymbol{p}_{i+1}}\cdot\boldsymbol{n}_{i-1}\label{eq_xy}\\
    \theta_{i} = \arccos(\frac{\overrightarrow{\boldsymbol{p}_{i}\boldsymbol{p}_{i-1}}\cdot\textnormal{Pj}_{\Sigma_{B_{i-1}}^{(xy)}}(\overrightarrow{\boldsymbol{p}_{i}\boldsymbol{p}_{i+1}})}{\|\overrightarrow{\boldsymbol{p}_{i}\boldsymbol{p}_{i-1}}\|\cdot\|\textnormal{Pj}_{\Sigma_{B_{i-1}}^{(xy)}}(\overrightarrow{\boldsymbol{p}_{i}\boldsymbol{p}_{i+1}})\|})\\
    \boldsymbol{q}_{i} = \boldsymbol{p}_{i}-r_c\cdot\tan{\dfrac{\theta_i}{2}}\cdot\dfrac{\overrightarrow{\boldsymbol{p}_{i-1}\boldsymbol{p}_{i}}}{\|\overrightarrow{\boldsymbol{p}_{i-1}\boldsymbol{p}_{i}}\|}\\
    \textnormal{Pj}_{\Sigma_{B_{i-1}}^{(yz)}}(\overrightarrow{\boldsymbol{p}_{i}\boldsymbol{p}_{i+1}})=\overrightarrow{\boldsymbol{p}_{i}\boldsymbol{p}_{i+1}}-\overrightarrow{\boldsymbol{p}_{i}\boldsymbol{p}_{i+1}}\cdot\dfrac{\overrightarrow{\boldsymbol{p}_{i-1}\boldsymbol{p}_{i}}}{\|\overrightarrow{\boldsymbol{p}_{i-1}\boldsymbol{p}_{i}}\|}\label{eq_yz}\\
    \alpha_{i} = \arccos(\dfrac{\boldsymbol{n}_{i-1}\times\overrightarrow{\boldsymbol{p}_{i-1}\boldsymbol{p}_{i}}}{\|\overrightarrow{\boldsymbol{p}_{i-1}\boldsymbol{p}_{i}}\|}\cdot\frac{\textnormal{Pj}_{\Sigma_{B_{i-1}}^{(yz)}}(\overrightarrow{\boldsymbol{p}_{i}\boldsymbol{p}_{i+1}})}{\|\textnormal{Pj}_{\Sigma_{B_{i-1}}^{(yz)}}(\overrightarrow{\boldsymbol{p}_{i}\boldsymbol{p}_{i+1}})\|})\\
    \textnormal{Pj}_{\Sigma_{B_{i-1}}^{(xz)}}(\overrightarrow{\boldsymbol{p}_{i}\boldsymbol{p}_{i+1}})=\overrightarrow{\boldsymbol{p}_{i}\boldsymbol{p}_{i+1}}-\overrightarrow{\boldsymbol{p}_{i}\boldsymbol{p}_{i+1}}\cdot\dfrac{\boldsymbol{n}_{i-1}\times\overrightarrow{\boldsymbol{p}_{i-1}\boldsymbol{p}_{i}}}{\|\overrightarrow{\boldsymbol{p}_{i-1}\boldsymbol{p}_{i}}\|}\label{eq_xz}\\
    \beta_{i} = \arccos(\dfrac{\overrightarrow{\boldsymbol{p}_{i-1}\boldsymbol{p}_{i}}\cdot\textnormal{Pj}_{\Sigma_{B_{i-1}}^{(xz)}}(\overrightarrow{\boldsymbol{p}_{i}\boldsymbol{p}_{i+1}})}{\|\overrightarrow{\boldsymbol{p}_{i-1}\boldsymbol{p}_{i}}\|\cdot\|\textnormal{Pj}_{\Sigma_{B_{i-1}}^{(xz)}}(\overrightarrow{\boldsymbol{p}_{i}\boldsymbol{p}_{i+1}})\|})
\end{align}

Here, $\textnormal{Pj}_{A}(B)$ means projecting $B$ onto $A$. The $A$ is replaced by $\Sigma_{B_{i-1}}^{(xy)}$, $\Sigma_{B_{i-1}}^{(yz)}$, $\Sigma_{B_{i-1}}^{(xz)}$ in equations \eqref{eq_xy}, \eqref{eq_yz}, and \eqref{eq_xz} respectively. They indicate projecting on to the $xy$, $yz$, $xz$ planes of the $\Sigma_{B_{i-1}}$ coordinate system.

\section{Combined Task and Motion Planning Considering Pruning}
\label{sec:tamp}
\begin{figure}[!tpb]
  \begin{center}
  \includegraphics[width=.95\linewidth]{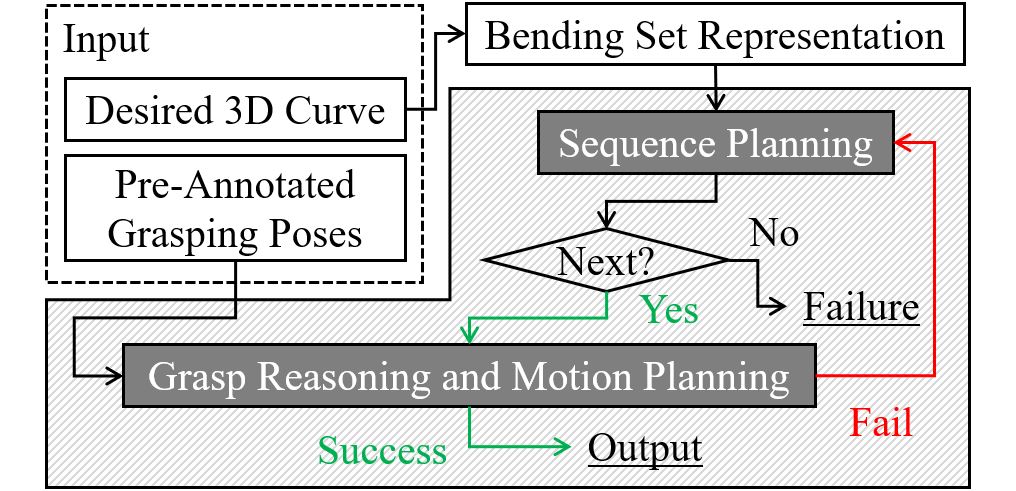}
  \caption{Workflow of the proposed planner. The gray region is the combined task and motion planning component.}
  \label{fig:workflow}
  \end{center}
\end{figure}

The overview workflow of our planner, with a particular focus on the combined task and motion planning part, is shown in Fig. \ref{fig:workflow}. First, it takes the desired shape as input and down-sample it to many key points. Then it generates a bending sequence considering the collision constraints by incrementally permuting the bending order and pruning the searching tree. Finally, together with the pre-annotated grasp posed of metal wire primitive, the workflow will produce a sequence of kinematic metal wire poses and send them to the motion planner to determine grasp poses and plan manipulation motion. It will iterate through the pre-annotated grasp poses to find the candidates that can finish all the motions. The essential combined task and motion planning component is illustrated in the gray region. Its details will be presented in the following subsections.

\subsection{Bending Sequence Planning}
\label{sec:bsp}
A feasible bending sequence should fulfill the following constraints: (i) The wire does not collide with the machine or any other obstacles before and after bending. (ii) The flange is long enough to contact the die roller and punch roller before and after bending. If the metal wire is unable to touch either of the rollers, the bending candidate $B_{i}$ pertaining to the current shape of the metal wire will be reported as infeasible as shown in Fig.\ref{fig:working_range}(c). Determining such a sequence is difficult as a set of $n$ bends can be arranged in $n!$ ways. Thus, instead of brutally evaluating all permutations, we leverage pruning to accelerate searching speed. The algorithm is shown in Algorithm \ref{alg:prune}. The sequence is initialized with $s$=$[0,1,...,n]$. The algorithm will explore the bending candidates one by one. If collision or infeasible bending appears at $B_{j}$, the remaining actions in the sequence no longer need to be explored. The failed sequence $\boldsymbol{s}[:j]$ will be stored in a tree structure $\boldsymbol{t}$. A new sequence not in $\boldsymbol{t}$ will be generated by Depth First Search (DFS), and the next iteration will be examined, and failures will be stored similarly until a feasible sequence is found. The time efficiency of the pruned search will be detailed in Section \ref{sec:exp}.
\begin{algorithm}[!tbp]
    \DontPrintSemicolon
    \SetKwInput{KwInput}{Input}                % Set the Input
    \SetKwInput{KwOutput}{Output}              % set the Output
    \KwInput{ 
        $\boldsymbol{B}={B_{0},B_{1}...,B_{n}}$, a bending set;\newline
        $\boldsymbol{t}$, invalid sequence tree;\newline
        $\boldsymbol{s}=[0,1,..,n]$, a bending sequence;\\
        }
    \Begin{
        \While{$\boldsymbol{s}$ is not empty}{
            \uIf{$\boldsymbol{s}$ is feasible}{
                Grasp reasoning and motion planning\\
                \lIf{success} {
                    \Return{$\boldsymbol{s}$}
                }
            }
            $\boldsymbol{t}\leftarrow$ update($\boldsymbol{t}$, $\boldsymbol{s}$)\\
            $\boldsymbol{s}\leftarrow$ Find a new sequence not in $\boldsymbol{t}$\\
        }
        \Return False
    }
    \caption{Prune Search} 
    \label{alg:prune} 
\end{algorithm}

\subsection{Grasp Reasoning and Motion Planning}
\begin{figure}[!tpb]
  \begin{center}
  \includegraphics[width=\linewidth]{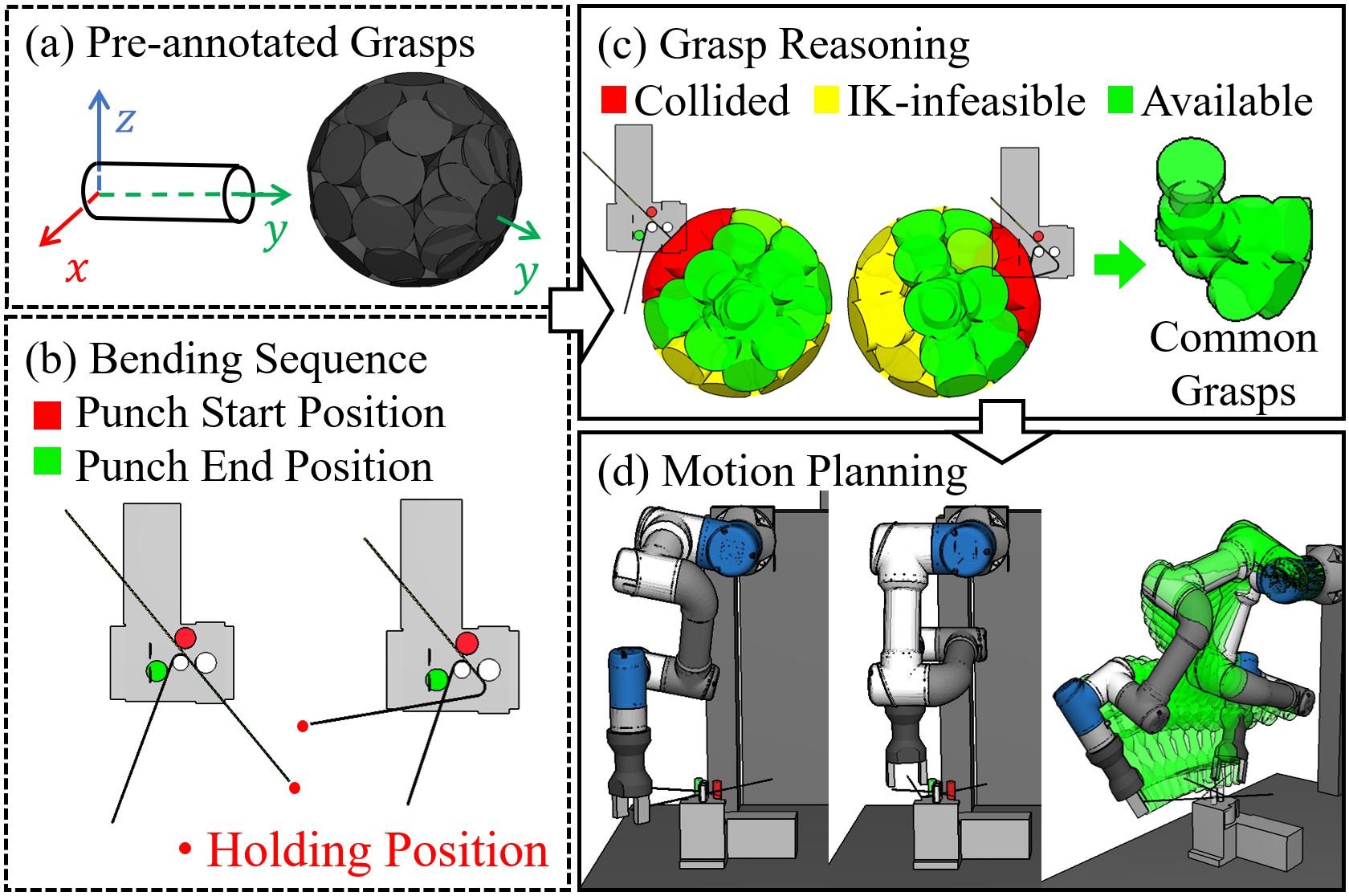}
  \caption{Workflow of grasp reasoning and motion planning. The inputs are shown in dashed boxes on the left. The results are in the solid boxes on the right. (a) Pre-annotated grasps. (b) Planned bending sequence. (c) Example of grasp reasoning. (d) Example of motion planning result.}
  \label{fig:gp_mp}
  \end{center}
\end{figure}

This subsection uses a simple example with two bends to present the grasp reasoning and motion planning. The input of this step is: (1) Pre-annotated grasp poses for the metal wire; (2) Wire poses generated by sequence planning. The input will be used to reason IK-feasible and collision-free grasps. Fig. \ref{fig:gp_mp}(a) shows an example of grasp reasoning. The collided grasps are shown in red, the IK-infeasible grasps are shown in yellow, and the available grasps are shown in green. The grasps available for all bends are named ``common grasps''. A robot can hold the wire using a ``common grasp'' at one pose and move it to another without regrasp. After obtaining the common grasps, the motion planning will iterate through them to plan the joint space motion for moving a wire to the pre-defined poses. Fig. \ref{fig:gp_mp}(b) exemplifies a planned motion. 

\section{Experiments and Analysis}
\label{sec:exp}
\begin{figure}[!tbp]
    \centering
    \includegraphics[width=\linewidth]{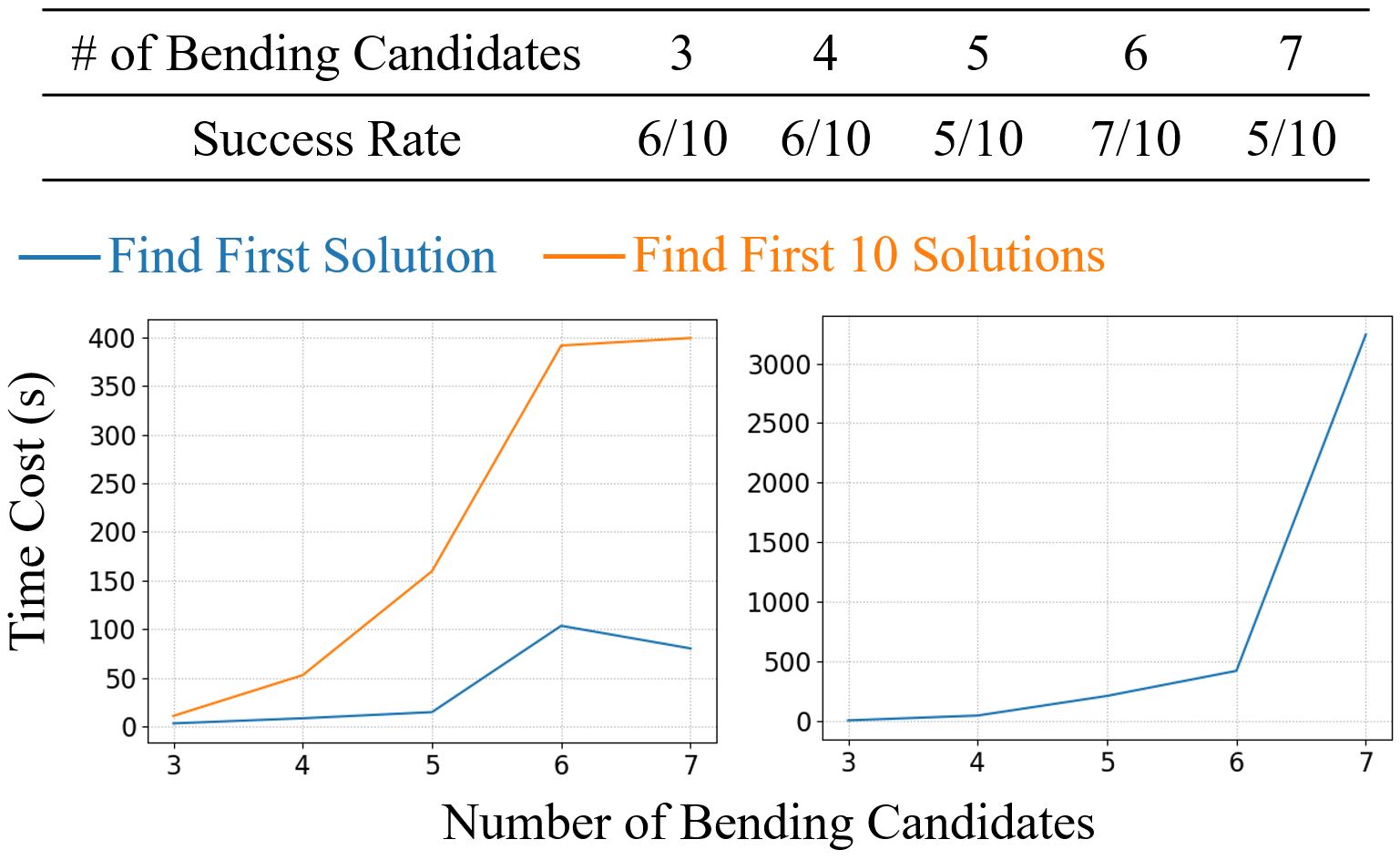}
    \caption{Statistical performance of sequence planning. Upper Table: Success rates under different number of bending candidates. Lower Graphs: Time costs.}
    \label{fig:seq_exp}
\end{figure}
\begin{figure}[!tbp]
    \centering
    \includegraphics[width=\linewidth]{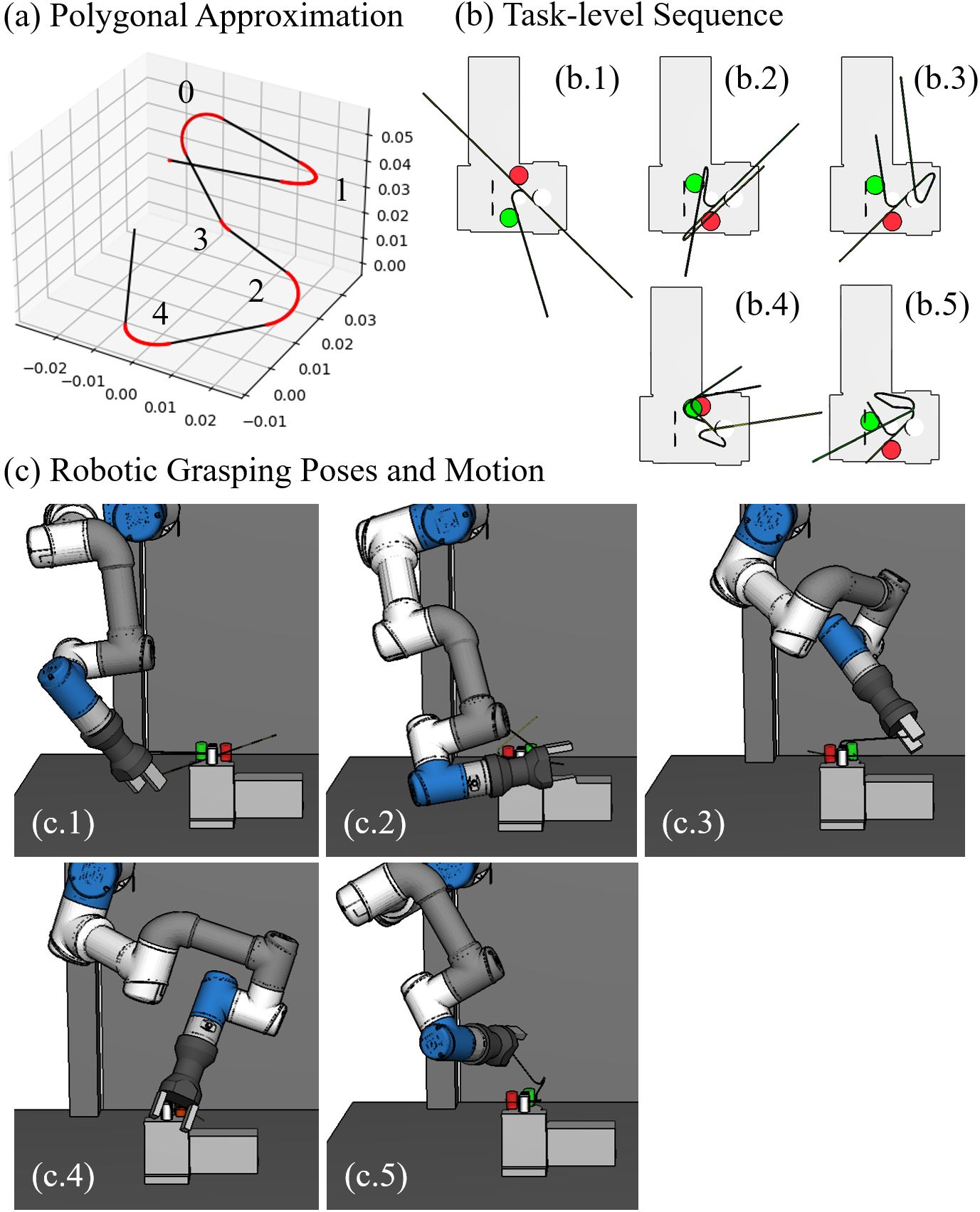}
    \caption{One exemplary result planned using our method.}
    \label{fig:seq_exp2}
\end{figure}
\begin{figure*}[]
    \centering
    \includegraphics[width=\linewidth]{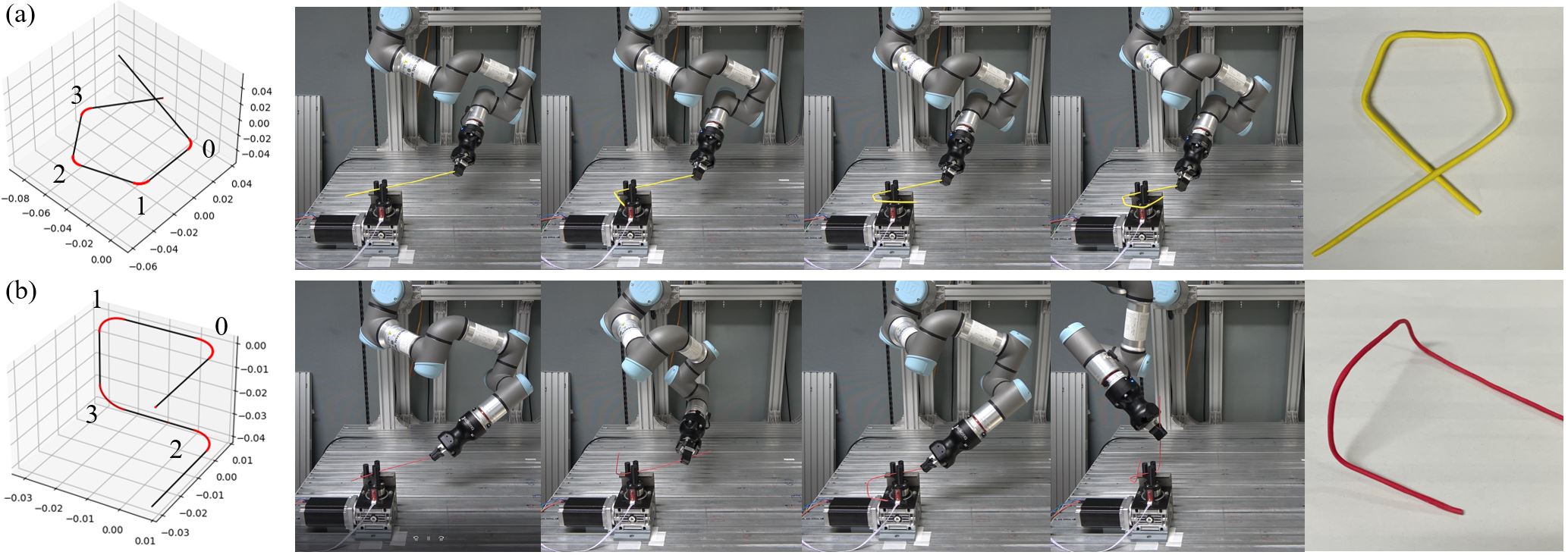}
    \caption{Experiment of two polygon bending task. The goal shape is shown in the first column, and the planned bending sequence is annotate on it. Following are the corresponding robot motion and final result.}
    \label{fig:exp}
\end{figure*}
Our experimental platform includes a PC with an Intel Core i7-10700 CPU and 32 GB memory, and a UR3e robotic arm equipped with a Robotiq Hand-E two-finger parallel gripper. A prototype of the bending machine was previously shown in Fig. \ref{fig:teaser}. It is fixed on a work table under an UR3e collaborative robotic arm (Max payload: 3kg).

\subsection{Sequence Planning}

In this subsection, we demonstrate some planning results and examine the time efficiency of the sequence planner. The experiments were carried out by randomly generating bending sets and planning their order using the developed method. Especially, we randomized the parameters bending angle $\theta_{i} \in [\pi,\pi]$, twisting angle $\alpha_{i} \in [-\pi,\pi]$ and lift angle $\beta_{i} \in [-\pi/18,\pi/18]$. The start position of each bending candidate, $q_{i}$, was set at a random position that did not overlap with others. Five groups of curves with the number of bending candidates in approximated bending sets ranging from 3 to 8 were evaluated. Each group included ten random desired curves. 

Fig. \ref{fig:seq_exp} shows the detailed statistic performance. The upper table shows the success rate of our planner with a changing number of bending candidates. The lower two graphs show the time cost of success and failure cases separately. The time cost increases with the number of bends, and this tendency is significant in failed cases.

Fig. \ref{fig:seq_exp2} shows the result of a 3D shape with five bending candidates. The approximated bending set is illustrated in (a). One feasible bending sequence is illustrated in (b). Key poses of a robot action sequence are illustrated in (c). 

\subsection{Performance of Real-World Executions}
We also carried out real-world experiments to check the final resultant bending. The goal here is to compare the desired shape and the bending result. The desired shapes include a 2D and a 3D polygon with a side length of 50mm. The metal wire used is galvanized steel wire with 1.6mm (red) and 2.6mm (yellow) diameters. Fig. \ref{fig:exp} shows the real-world results. The goal shapes are visualized in the first column. The planned robot motion and final result are shown on the left. The details of executions are shown in a supplementary video.

\section{Conclusions and Future Work}
We presented a combined task and motion planning-based planner for a robot arm bend metal wire by collaborating with a bending machine. It enables a low payload robot arm to curve a metal wire with high stiffness. Currently, we treat the planning problem hierarchically and solve each sub-problem using a specially designed planner. In the future, we are interested in leveraging reinforcement learning to solve them together in a single shot to improve efficiency.

\bibliographystyle{IEEEtran}
\bibliography{citations}

\end{document}